\documentclass[conference]{IEEEtran}
\IEEEoverridecommandlockouts        
                                    
\usepackage{graphicx}
\usepackage{svg}

\usepackage{dtklogos}
\usepackage{cite}
\usepackage{graphicx}
\usepackage{caption}
\usepackage{booktabs}

\usepackage{float}
\usepackage{mathtools, cases}
\usepackage{amsmath,empheq}
\usepackage{eqparbox} 
\usepackage{bm} 
\usepackage{multirow}
\usepackage{lipsum} 

\begin{document}
\title{Deep Network for Capacitive ECG Denoising}
\author{
\IEEEauthorblockN{Vignesh Ravichandran and \\ Balamurali Murugesan}
\IEEEauthorblockA{Healthcare Technology Innovation Centre \\Indian Institute of Technology Madras \\
Chennai, India \\
vignesh.ravi@htic.iitm.ac.in\\}
\\\\
\IEEEauthorblockN{Keerthi Ram, Preejith S.P and \\ Jayaraj Joseph} 
\IEEEauthorblockA{Healthcare Technology Innovation Centre \\Indian Institute of Technology Madras \\
Chennai, India}
\\\vspace*{-1cm}
\and
\IEEEauthorblockN{Sharath M Shankaranarayana}
\IEEEauthorblockA{Department of Electrical Engineering \\Indian Institute of Technology Madras \\
Chennai, India\\}
\\\\\\\\\
\IEEEauthorblockN{Mohanasankar Sivaprakasam}
\IEEEauthorblockA{Department of Electrical Engineering \\Indian Institute of Technology Madras \\ Healthcare Technology Innovation Centre\\
Chennai, India}
\\\vspace*{-1cm}
}
\maketitle

\begin{abstract}
Continuous monitoring of cardiac health under free living condition is crucial to provide effective care for patients undergoing post operative recovery and individuals with high cardiac risk like the elderly. Capacitive Electrocardiogram (cECG) is one such technology which allows comfortable and long term monitoring through its ability to measure biopotential in conditions without having skin contact. cECG monitoring can be done using many household objects like chairs, beds and even car seats allowing for seamless monitoring of individuals. This method is unfortunately highly susceptible to motion artifacts which greatly limits its usage in clinical practice. The current use of cECG systems has been limited to performing rhythmic analysis. In this paper we propose a novel end-to-end deep learning architecture to perform the task of denoising capacitive ECG. The proposed network is trained using motion corrupted three channel cECG and a reference LEAD I ECG collected on individuals while driving a car. Further, we also propose a novel joint loss function to apply loss on both signal and frequency domain. We conduct extensive rhythmic analysis on the model predictions and the ground truth. We further evaluate the signal denoising using Mean Square Error(MSE) and Cross Correlation between model predictions and ground truth. We report MSE of 0.167 and Cross Correlation of 0.476. The reported results highlight the feasibility of performing morphological analysis using the filtered cECG. The proposed approach can allow for continuous and comprehensive monitoring of the individuals in free living conditions.
\end{abstract}
\def\IEEEkeywordsname{Keywords}

\section{INTRODUCTION}

The increasing prevalence of cardiovascular diseases worldwide warrants the need to develop efficient tools for early screening cardiac ailments. Electrocardiogram (ECG) is a biosignal which is clinically used to diagnose a wide range of cardiovascular diseases. Conventional ECG recording systems use Ag/AgCl electrodes to pick up the biopotential associated with the heart activity through direct contact with the skin. This method, however, is unsuited for long term monitoring usage due to the discomfort and skin irritation associated with gel type electrodes. The use of dry conformal electrodes too introduces disturbances to the daily activity of the users due to the use of lead cables. Capacitive ECG provides an alternative for non-contact measurement of cardiac biopotential allowing for a viable compromise between user comfort and ubiquitous cardiac monitoring. The measurement is carried out using noncontact electrodes that sense the cardiac bioelectric signal through any insulator medium between the skin and the electrode \cite{komensky2012ultra}. The insulator medium can either be clothing, hair or air. cECG has been evaluated in several applications ranging from its usage in a common chair \cite{aleksandrowicz2007wireless} to sensors below the bed \cite{lim2007ecg} to driver monitoring in automobiles \cite{leonhardt2008non}. The value of cECG arises from its potential to detect cardiac anomalies such as arrhythmias and provide clinically useful parameters like heart rate variability. This, however, is held back due to its high sensitivity to motion artifacts \cite{sun2016capacitive}. Motion artifacts has a significant presence in cECG in applications where the user is conscious like while driving an automobile seat or while sitting on a chair. The motion artifacts are pronounced in cECG when compared to contact based ECG methods due to the variation of the coupling capacitance caused by relative motion of the user to the electrode \cite{ottenbacher2009motion}. This greatly increases the number of false detections in traditional QRS detectors. A few approaches have been proposed to reduce the effect of motion artifacts in capacitive ECG. Eilebrecht \textit{et al.} proposed using data from the accelerometer to perform adaptive filtering \cite{eilebrecht2011motion}. Choi \textit{et al.} proposed using an additional reference capacitive electrode to profile the nature of the motion artifact before applying an adaptive filter \cite{choi2016reduction}. Recently, Christoph \textit{et al.}  \cite{HoogAntink2018} have proposed using a deep learning network to perform the task of peak detection in a three channel cECG data collected under significant motion artifacts from subjects driving a car. The approach used three convolutional channels along with a moving window to detect peak location. This method, however, only provides peak locations as opposed to denoising the signal. Deriving morphological information from cECG is crucial for detecting a wide range of cardiac diseases. The cECG denoising application is analogous to the image denoising where certain filters need to be applied to the noisy image. With the advent of deep learning, however, the task of image denoising has been carried out using autoencoders \cite{vincent2010stacked}. One specific architecture, Unet has provided significant results while denoising CT images \cite{gupta2018cnn}. Stoller \textit{et al.}  proposed WaveU-net to perform sound source separation in the 1D audio domain using Unet \cite{Stoller2018}. Similarly, performing denoising on cECG would extend its clinical utility from rhythmic analysis to providing morphological analysis for screening cardiac ailments in free living conditions.
\par    
The contributions of this paper are as follows:
\begin{itemize}
\item {We propose a novel end-to-end deep learning network to perform capacitive ECG denoising from multichannel input capacitive ECG.}
\item{We propose a novel joint loss function which combines L1 loss between model prediction and ground truth on the signal domain with Fourier domain L1 loss between model prediction and ground truth.}
\item{We study the performance of the proposed network using the UnoViS\_auto2012 dataset for the task of localizing R-peaks using a comprehensive HRV analysis.}
\item{We evaluate the feasibility of deriving morphological significance from the ECG generated by the proposed model by comparing similitude and signal reconstruction error against a reference LEAD I ECG.}
\end{itemize}

\section{METHODOLOGY}
\subsection{Problem formulation}
The task of the proposed deep learning model is to extract a denoised LEAD I ECG from given multichannel capacitive ECG signals. The dataset $X=\{(x^{(1)},y^{(1)}),(x^{(2)},y^{(2)}),....,(x^{(m)},y^{(m)})\}$ consists of the multichannel capacitive ECG signal $x^{(i)} = x^{(i)}_{ch1},x^{(i)}_{ch2},x^{(i)}_{ch3}$ and reference LEAD I ECG signal $y^{(i)}$, where $x^{(i)}_{ch1},x^{(i)}_{ch2},x^{(i)}_{ch3} \in R^{n}$ and $y^{(i)} \in R^{n}$.

The proposed deep learning network is designed in the topology of an encoder-decoder network making use of fully convolutional blocks. The encoder section produces feature vectors $z_{1}^{(i)}$ by performing repeated downsampling of the input $x^{(i)}$. The decoder section however, performs repeated upsampling using $z_{1}^{(i)}$ as its input and produces the predicted LEAD I ECG signal $y_{pred}^{(i)}$. These are represented by equations \ref{eq:encoder} and \ref{eq:decoder}.

\begin{equation}\label{eq:encoder}
z_{1}^{(i)} = F_{1}(x^{(i)};\theta_{1})
\end{equation}
\begin{equation}\label{eq:decoder}
y_{pred}^{(i)} = F_{2}(z_{1}^{(i)};\theta_{2})
\end{equation} Where $F_{1}$, $F_{2}$ are function representations of the encoder and decoder with parameters $\theta_{1}$  and $\theta_{2}$. The decoder network outputs $y_{pred}^{(i)}$ which denotes the predicted LEAD I ECG.

\begin{figure*}

  \includegraphics[width=\textwidth]{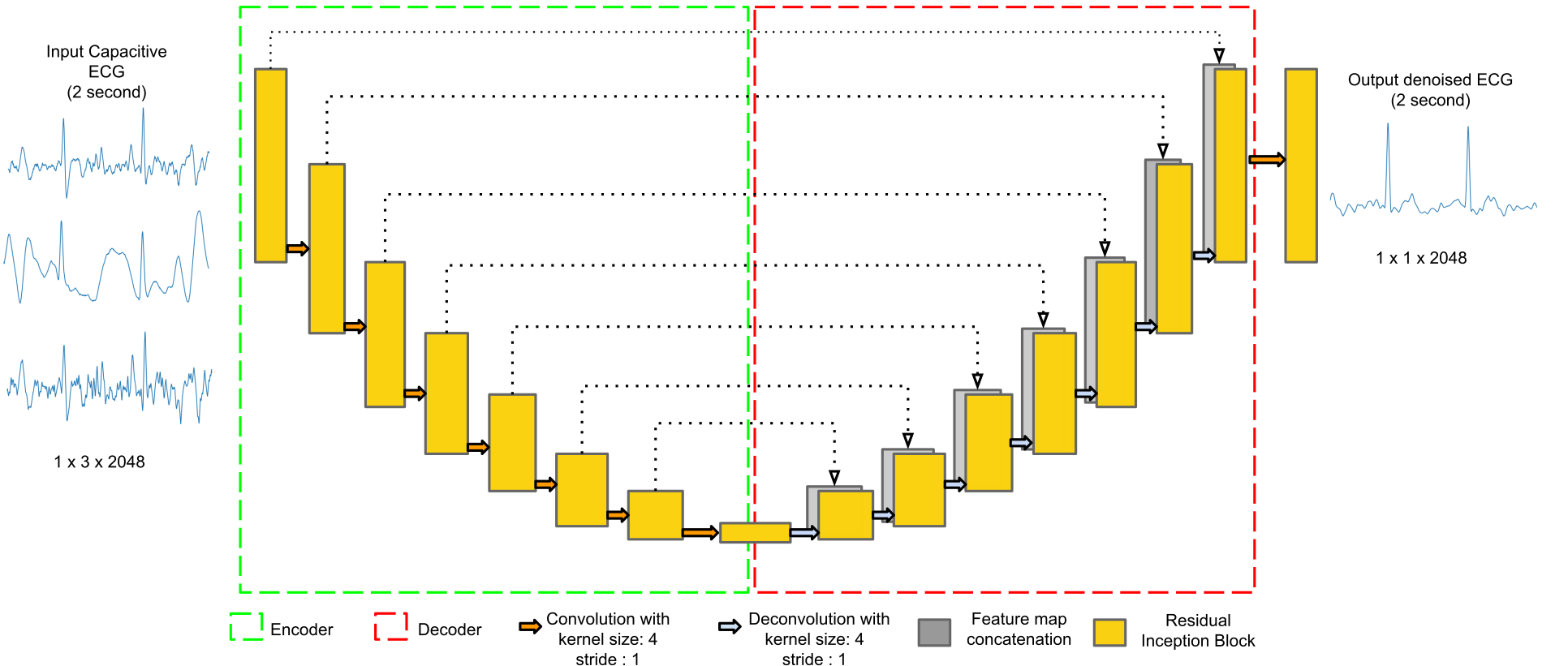}
  \caption{Proposed CNN Architecture}
  \label{arch}
\end{figure*}


\subsection{Network Architecture}

The proposed network was inspired from the IncResU-Net network which is used for 2D medical image segmentation application \cite{ShankaranarayanaJBHI}. The network architecture of the proposed fully convolutional network for performing capacitive ECG denoising can be seen in Fig. \ref{arch}. The network architecture consists of two sections namely the encoder and decoder. The downsampling operation is performed in encoder section with the 3 channel input cECG signals. The encoder section is made of 8 levels, at each level convolution operation is performed using filters of size 1x4. This is then followed by Batch Normalization \cite{ioffe2015batch}, leaky ReLU \cite{xu2015empirical} with slope 0.2, Dropout and finally dilated residual inception block. Batch Normalization helps in accelerating the convergence of model as it transforms any input data to a distribution with zero mean and unit variance. Non-linearity in the network is achieved by introducing the leaky ReLU activation function. Dropout helps in reducing interdependent learning amongst neurons by randomly dropping out nodes in the network connection. Dilated residual inception blocks provide a larger receptive field without a significant increase in the number of parameters of the model. Applying multiple convolutions at different dilation rates within the dilated block prevents the vanishing gradient problem and significantly reduces the convergence time during training. In order to improve training efficiency during downsampling, strided convolution operation was used instead of Max-pooling to preserve spatial information \cite{springenberg2014striving}. The input size is decreased during downsampling operation while increasing the number of filters at each convolution layer. The number of filters in the encoder section are incremented by a factor of two till the total filters in a level is 512. The total filters in subsequent levels in the encoder are maintained at 512. Finally a 1x1 1D convolution operation was performed on 8$^{th}$ level of the encoder. Upsampling is carried out in the decoder section using the deconvolution operation at each level similar to encoder section. At each level, features from decoder section were concatenated with corresponding encoder pair similar to the original U-Net. In the final level of the decoder, a 1$\times$1  1D  convolution operation is used to obtain the desired output LEAD I ECG predictions from the feature maps.

\subsection{Loss Function}

In the domain of deep learning based Magnetic Resonance Imaging reconstruction, it was found that incorporating loss terms from both image domain and frequency domain between the ground truth and model prediction resulted in better reconstruction \cite{yang2018dagan}. A similar approach can be used in the task of capacitive ECG denoising, where it is crucial to enforce similarity in both the signal and frequency domains.  Hence, for training the proposed model, we apply a joint loss comprised of both signal domain loss and frequency domain loss which are represented as

\begin{equation}
\begin{aligned}
L_{s}(\theta)   = \sum_{i=1}^{m}SmoothL_1(y^{(i)}-y_{pred}^{(i)})
\end{aligned}
\end{equation}

\begin{equation}
L_{f}(\theta) = \sum_{i=1}^{m}SmoothL_1(Y^{(i)}-Y_{pred}^{(i)})
\end{equation}

\begin{equation}
L_{total}(\theta) = \alpha L_{s}(\theta) + \beta L_{f}(\theta)
\end{equation}


\begin{equation}
  SmoothL_1(y_{diff}) = 
  \begin{cases}
    0.5(y_{diff})^2 &\text{if} \, \lvert y_{diff} \rvert < 0 \\
     \lvert y_{diff}\rvert -0.5 &\text{otherwise,}
  \end{cases}
\end{equation}

Where $L_{s}$ and $L_{f}$ correspond to the signal domain and frequency domain
losses. $y^{(i)}$ and $y_{pred}^{(i)}$ refer to the LEAD I ECG ground truth and model prediction, $Y^{(i)}$ and $Y_{pred}^{(i)}$ are the corresponding fourier tranform of the signals $y^{(i)}$ and $y_{pred}^{(i)}$. The two losses are added together after scaling with coefficients $\alpha$ and $\beta$ resulting in total loss, $L_{total}$. $\theta$ refers to the parameters of the proposed network which require optimization.

\subsection{Training Method}

During training, the network was initialized using random weights.
The signal domain loss term, as defined in equation 3, was found between the network prediction and ground truth. The frequency domain loss term as defined in equation 4 is found by applying 1024 point Fast Fourier Transform (FFT) for the network prediction and ground truth and determining the $SmoothL1$ loss between the two. The network parameters were optimized using Stochastic Gradient Descent for a minibatch comprising of 256 input windows. The learning rate of the network was set to 0.01 and the momentum to 0.7. The training was carried out for 2500 epochs. The model was developed and implemented in PyTorch \cite{paszke2017automatic}. The training was carried out in a workstation using a i7 8700K CPU and Nvidia GTX1080Ti 11GB GPU.

\section{DATASET DESCRIPTION}
The UnoViS database is a comprehensive database for research in the domain of unobtrusive medical monitoring through measurement of PPG and capacitive ECG in free living conditions along with reference ECG \cite{wartzek2015unovis}. The database includes recording in diverse conditions like while driving a car and while sleeping. The data collected from the car driving condition, referred to as the UnoViS$\_$auto2012 dataset is the largest recording, accounting for over 13 hours. The UnoViS$\_$auto2012 dataset is however highly influenced by motion artifacts compared to the other conditions. The dataset is composed of 31 recording sessions of 6 different subjects whilst driving in different road types like highways, curvy roads and bad roads. Three bipolar capacitive ECG leads were obtained similar to the Einthoven's triangle. Simultaneous reference LEAD I ECG was obtained from a clinical grade ICU monitor. Out of the 31 recordings, two recordings were sampled at 200 Hz while the remaining datasets were sampled at 1000 Hz. Both the capacitive ECG signals and the reference LEAD I ECG were resampled to 1024 Hz to ensure compatibility with the proposed network which requires an input and reference of size 2048 (1024$\times$2). We create a train-test split by using 22 files for training and 7 files for testing. Two second length windows of three lead capacitive ECG signals and reference ECG signal are extracted from both the train and test files. Table \ref{table_dataset1} summarizes the dataset size used for training and testing.

\begin{table}[]
\caption{Dataset Description}
\label{table_dataset1}
\resizebox{0.47\textwidth}{!}{%
\begin{tabular}{|c|c|c|c|c|c|}
\hline
\textbf{Dataset Name}       & \textbf{Sampling} & \textbf{Input signals}                                             & \textbf{Reference signals}                                     & \begin{tabular}[c]{@{}c@{}}\textbf{Train-set} \\ \textbf{windows}\end{tabular} & \begin{tabular}[c]{@{}c@{}}\textbf{Test-set}\\ \textbf{windows}\end{tabular} \\ \hline
UnoViS$\_$auto2012 & 1000 Hz  & \begin{tabular}[c]{@{}c@{}}3 x capacitive ECG \\ LEAD's\end{tabular} & \begin{tabular}[c]{@{}c@{}}LEAD I \\ ECG\end{tabular} & 11359                                                        & 2938                                                       \\ \hline
\end{tabular}
}
\end{table}

\section{EXPERIMENTS AND RESULTS}

\begin{table*}[!htb]
\centering
\caption{HRV analysis of proposed models and ground truth ECG}
\label{hrv}
\resizebox{0.98\textwidth}{!}{%
\begin{tabular}{|c|c|c|c|c|c|c|c|c|c|c|c|c|c|c|}
\hline
\multirow{2}{*}{\textbf{\begin{tabular}[c]{@{}c@{}}\\File \\ Number\end{tabular}}} & \multicolumn{4}{c|}{\textbf{\begin{tabular}[c]{@{}c@{}}Ground Truth ECG\\ HRV metrics\end{tabular}}} & \multicolumn{5}{c|}{\textbf{\begin{tabular}[c]{@{}c@{}}Model predictions (L1)\\ HRV metrics\end{tabular}}} & \multicolumn{5}{c|}{\textbf{\begin{tabular}[c]{@{}c@{}}Model predictions (L1+RFFT)\\ HRV metrics\end{tabular}}} \\ \cline{2-15} 
 & \textbf{\begin{tabular}[c]{@{}c@{}}Mean RR \\ intervals\\ (s)\end{tabular}} & \textbf{\begin{tabular}[c]{@{}c@{}}RMSSD\\ (s)\end{tabular}} & \textbf{\begin{tabular}[c]{@{}c@{}}pNN50 \\ (\%)\end{tabular}} & \textbf{LF/HF} & \textbf{\begin{tabular}[c]{@{}c@{}}Mean \\ RR\\ intervals\\  (s)\end{tabular}} & \textbf{\begin{tabular}[c]{@{}c@{}}RMSSD \\ (s)\end{tabular}} & \textbf{\begin{tabular}[c]{@{}c@{}}pNN50 \\ (\%)\end{tabular}} & \textbf{LF/HF} & \textbf{\begin{tabular}[c]{@{}c@{}}Cross \\ Correlation\end{tabular}} & \textbf{\begin{tabular}[c]{@{}c@{}}Mean RR \\ intervals\\ (s)\end{tabular}} & \textbf{\begin{tabular}[c]{@{}c@{}}RMSSD\\ (s)\end{tabular}} & \textbf{\begin{tabular}[c]{@{}c@{}}pNN50 \\ (\%)\end{tabular}} & \textbf{LF/HF} & \textbf{\begin{tabular}[c]{@{}c@{}}Cross \\ Correlation\end{tabular}} \\ \hline
1 & 0.742 & 0.077 & 40.10\% & 0.563 & 0.785 & 0.089 & 47.20\% & 1.221 & 0.248 & 0.821 & 0.084 & 44.70\% & 1.343 & 0.458 \\ \hline
2 & 0.782 & 0.027 & 6.42\% & 5.271 & 0.765 & 0.039 & 13.41\% & 0.347 & 0.347 & 0.779 & 0.0369 & 9.24\% & 5.828 & 0.384 \\ \hline
3 & 0.785 & 0.0258 & 3.85\% & 5.326 & 0.695 & 0.082 & 17.26\% & 1.73 & 0.272 & 0.716 & 0.063 & 13.27\% & 2.791 & 0.463 \\ \hline
4 & 0.787 & 0.0253 & 3.92\% & 5.267 & 0.712 & 0.097 & 36.13\% & 2.617 & 0.316 & 0.726 & 0.067 & 24.63\% & 3.182 & 0.374 \\ \hline
5 & 0.815 & 0.034 & 4.37\% & 5.326 & 0.697 & 0.082 & 37.32\% & 0.831 & 0.461 & 0.728 & 0.054 & 7.36\% & 8.857 & 0.464 \\ \hline
6 & 0.772 & 0.025 & 4.58\% & 7.31 & 0.726 & 0.073 & 28.18\% & 3.184 & 0.273 & 0.755 & 0.033 & 8.95\% & 5.456 & 0.347 \\ \hline
7 & 0.769 & 0.0287 & 8.41\% & 3.718 & 0.734 & 0.027 & 7.40\% & 3.878 & 0.449 & 0.754 & 0.031 & 9.39\% & 3.611 & 0.464 \\ \hline
\end{tabular}
}
\end{table*}
Two different models are proposed for the task of cECG denoising, the first model was exclusively trained on the signal domain $SmoothL1$ loss while the second model was trained on both the signal domain and frequency domain $SmoothL1$ loss. The evaluation of the proposed models are carried out in two different methods. First comparison is performed on the RR interval and HRV related parameters derived from the denoised signals and reference signal. Second comparison is performed between the denoised predictions between the proposed models and the ground truth ECG. 

\subsection{R-peak detection evaluation}

cECG based Heart Rate, HRV and rhythm analysis have shown significant clinical value  \cite{arcelus2013design}\cite{oehler2009novel}. A common source of error in such cECG based rhythm metrics is motion induced artifacts. An important criteria for evaluating such denoising solutions is through rhythm analysis. The predictions from the proposed models along with the corresponding reference LEAD I ECG for the 7 test files are provided to a Hamilton QRS detector \cite{hamilton2003open} after applying a 4$^{th}$ order butteworth bandpass filter with a cutoff frequency 0.5-60 Hz. For each of the test files, RR intervals obtained from the QRS detector are provided to the Kubios HRV analysis tool for the model predictions and ground truth \cite{tarvainen2014kubios}. Metrics like Mean RR intervals, Root Mean Square of the Successive Differences (RMSSD), pNN50 and Low Frequency-High Frequency ratio (LF/HF) were obtained through Kubios HRV tool to provide a comprehensive understanding of the performance of the proposed model for enabling HR, HRV and rhythmic abnormality measurement. Further we obtain cross correlation metrics between the R location predictions of the models and the R location predictions from LEAD I ECG. Table \ref{hrv} shows HRV analysis of the proposed model and ground truth ECG.

\subsection{Denoising evaluation}

For this evaluation we apply Min-Max normalization to all the signals used for this comparison by scaling the values between the range 0 and 1. The MSE difference and cross-correlation metrics including lag between the model predictions and the reference LEAD I ECG signals are found for all the test set windows. Table \ref{compare} shows the  performance of both the model trained on signal domain $SmoothL1$ loss and the model trained on joint signal domain and frequency domain $SmoothL1$ loss using mean cross-correlation, mean lags and mean MSE. 

\begin{table}[]
\centering
\caption{Comparison for LEAD I ECG reconsturction}
\label{compare}
\begin{tabular}{|c|c|c|c|}
\hline
\textbf{Model} & \textbf{MSE} & \textbf{\begin{tabular}[c]{@{}c@{}}Cross \\ Correlation\end{tabular}} & \textbf{Lag} \\ \hline
L1 & 0.235 & 0.241 & 0.989 \\ \hline
L1 + RFFT & \textbf{0.167} & \textbf{0.476} & \textbf{0.998} \\ \hline
\end{tabular}
\end{table}

\section{DISCUSSION}
\begin{figure*}
  \includegraphics[width=\textwidth]{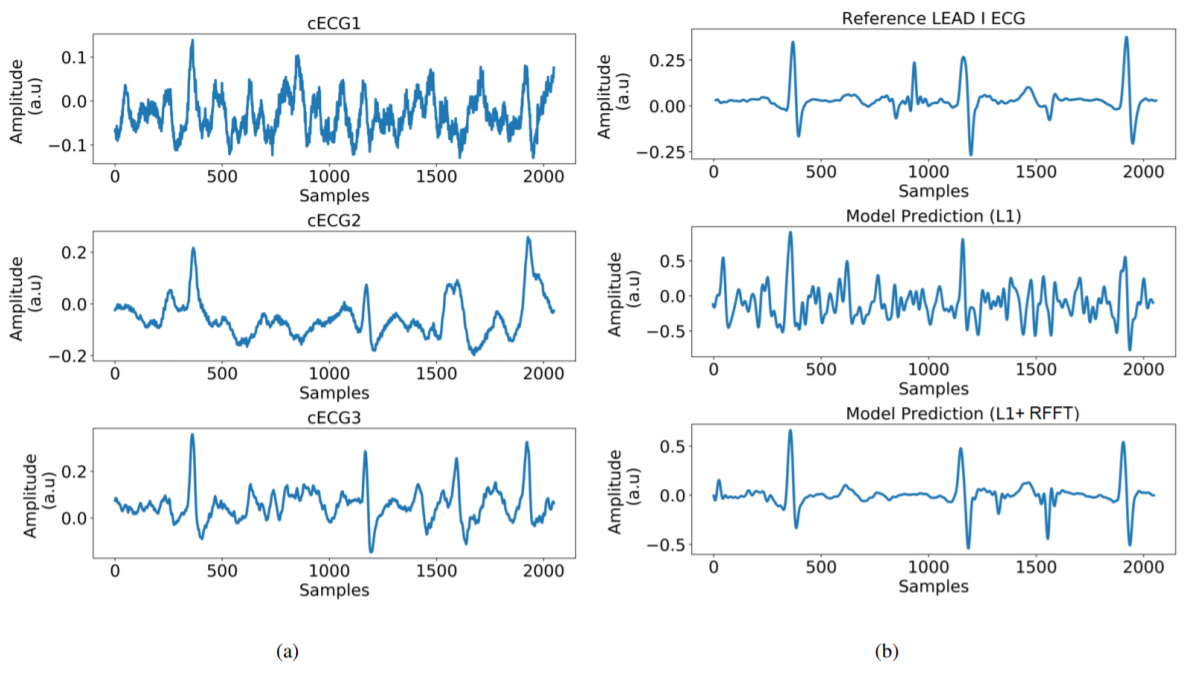}
\centering
\captionsetup{justification=centering}
\caption{(a) Sample input 3 channel capacitive ECG signal \\
(b) Reference Lead I ECG, Model predictions: Signal domain loss, Joint loss}
  \label{preds_new}
\end{figure*}

As seen in Table \ref{hrv}, the predictions of the proposed models provide similar R-peak localization metrics when compared with the ground truth R-peak locations. The model using the joint loss on both the signal domain and frequency domain in particular provides significant improvement over the model that was exclusively trained on the signal domain loss. The similarity of the HRV metrics derived from proposed models with that of the reference ECG can be observed. It indicates the utility of HRV measurement from the cECG predictions. From Table \ref{compare}, it can be seen that the proposed models, in particular the model trained with joint loss display high cross correlation and low MSE when compared with the reference LEAD I ECG. This showcases the feasibility of performing morphological analysis using the model predictions and cECG input signals. An example of a denoised cECG window can be seen in Fig. \ref{preds_new}. From the figure, it can be observed that the prediction of the model trained exclusively on the signal domain loss is vulnerable to artifacts whereas the prediction of the model trained on both the signal and frequency domain shows lower amount of noise.

  

\section{CONCLUSION}
The present work describes a novel approach to denoise multichannel cECG signals using a learning based model. The proposed end-to-end deep learning framework takes motion corrupted three channel cECG signals as input and LEAD I ECG as reference training. We further propose a novel joint training loss which applies loss on both the signal domain and frequency domain. Through comparison of the RR intervals of the reference LEAD I ECG and proposed model outputs we report high similarity in HRV analysis. The model trained on joint loss provides lower error when compared to the model trained exclusively on the signal domain loss. Further we study the feasibility of performing morphological analysis on the predicted ECG by obtaining signal similarity metrics for both variants of the model. With a mean correlation of 0.476 and MSE of 0.167 the proposed model shows the feasibility of conducting morphological analysis using the model predictions along with rhythm analysis. This can potentially allow for long term unobstrusive monitoring of ECG in free living conditions. Extensive validation is crucial to determine the capability of the proposed predictions for providing morphological information. Training has to be carried out on a wide range of cardiac anomalies along with a corresponding performance study. Future scope of the proposed study would be to collect more capacitive ECG data for abnormal condition. The development of a robust SQI metric is required to allow elimination of noisy predictions along with characterization of scaling variance in capacitive ECG associated with motion artifacts. A comprehensive study using an ECG segmentation algorithm is required to verify if the model predictions retain morphological integrity in different segments of ECG.

\section*{ACKNOWLEDGMENT}
We would like to acknowledge the Helmholtz-Institute for Biomedical Engineering who have provided the UnoVis capacitive ECG dataset.
\nocite{*}
\bibliographystyle{IEEEtran}
\bibliography{main}
\end{document}